\documentclass[sigconf]{acmart}
\AtBeginDocument{%
  \providecommand\BibTeX{{%
    \normalfont B\kern-0.5em{\scshape i\kern-0.25em b}\kern-0.8em\TeX}}}

\newcommand{\ap}{\textsc{SmartQuery}}
\usepackage{multirow}
\usepackage[inline]{enumitem}
\setcopyright{acmcopyright}

\copyrightyear{2022}
\acmYear{2022}
\setcopyright{acmcopyright}\acmConference[CIKM '22]{Proceedings of the 31st ACM International Conference on Information and Knowledge Management}{October 17--21, 2022}{Atlanta, GA, USA}
\acmBooktitle{Proceedings of the 31st ACM International Conference on Information and Knowledge Management (CIKM '22), Oct. 17--21, 2022, Atlanta, GA, USA}
\acmPrice{15.00}
\acmDOI{10.1145/3511808.3557701}
\acmISBN{978-1-4503-9236-5/22/10}

\begin{document}

\title{\ap{}: An Active Learning Framework for Graph Neural Networks through Hybrid Uncertainty Reduction}

\author{Xiaoting Li}
\email{xiaotili@visa.com}
\affiliation{%
  \institution{Visa Research}
  \city{Palo Alto}
  \state{CA}
  \country{USA}
}

\author{Yuhang Wu}
\email{yuhawu@visa.com}
\affiliation{%
  \institution{Visa Research}
  \city{Palo Alto}
  \state{CA}
  \country{USA}
}

\author{Vineeth Rakesh}
\email{vinmohan@visa.com}
\affiliation{%
  \institution{Visa Research}
  \city{Palo Alto}
  \state{CA}
  \country{USA}
}

\author{Yusan Lin}
\email{yusalin@visa.com}
\affiliation{%
  \institution{Visa Research}
  \city{Palo Alto}
  \state{CA}
  \country{USA}
}

\author{Hao Yang}
\email{haoyang@visa.com}
\affiliation{%
  \institution{Visa Research}
  \city{Palo Alto}
  \state{CA}
  \country{USA}
}

\author{Fei Wang}
\email{feiwang@visa.com}
\affiliation{%
  \institution{Visa Research}
  \city{Palo Alto}
  \state{CA}
  \country{USA}
}





\renewcommand{\shortauthors}{Xiaoting Li et al.}

\begin{abstract}

Graph neural networks have achieved significant success in representation learning. However, the performance gains come at a cost; acquiring comprehensive labeled data for training can be prohibitively expensive. Active learning mitigates this issue by searching the unexplored data space and prioritizing the selection of data to maximize model's performance gain. In this paper, we propose a novel method \ap{}, a framework to learn a graph neural network with very few labeled nodes using a hybrid uncertainty reduction function. This is achieved using two key steps: (a) design a multi-stage active graph learning framework by exploiting diverse explicit graph information and (b) introduce label propagation to efficiently exploit known labels to assess the implicit embedding information. Using a comprehensive set of experiments on three network datasets, we demonstrate the competitive performance of our method against state-of-the-arts on very few labeled data (up to $5$ labeled nodes per class).

\end{abstract}



\maketitle


\section{Introduction}
Graph neural networks (GNNs) have played a pivotal role in the success of graph representation learning where models can embed graph data into low-dimensional space \cite{perozzi2014deepwalk, tang2015line, wu2020comprehensive, xu2018powerful}. 
Specifically, GNNs follow a well-designed message-passing scheme, where the nodes aggregate and transform the information from their neighbors in each layer \cite{gilmer2017neural}. However, gathering enough labeled data for training can be expensive in many practical applications, e.g., financial risk modeling and corporate security analytics \cite{ding2020graph}. One can manually label the objects, but this quickly becomes impractical for large volumes of data. Selecting a part of data for labeling is also non-trivial as different sets of training nodes can lead to very different graph representation results \cite{gao2018active}. Consequentially, this challenges the feasibility of GNNs in real-world applications.

Active learning (AL) \cite{settles2009active, gal2017deep, abel2019regional, cai2017active, wu2019active, aggarwal2014active, gao2018active} is a popular technique to alleviate label sparsity issue. The general idea is to dynamically query the most informative instances from the unlabeled data. The queried data points are then labeled by an oracle and iteratively integrated to expand the pool of the training data \cite{hu2020graph}. The pool size is restrained by a given labeling budget. Traditional AL algorithms are designed for independent and identically distributed (i.i.d.) data such as natural language processing and computer vision. However, there are few studies that address the application of AL on graph-structured data where there are strong interactions between nodes.


Studies such as \cite{cai2017active, gao2018active, hu2020graph} and \cite{abel2019regional} integrate GNNs with AL to improve the representative power of graph embeddings. Some of the methods \cite{cai2017active, gao2018active} combine AL metrics that work independently on different input spaces to compute the data informativeness and fail to interactively capture the knowledge. While \cite{hu2020graph} requires full labels of source graphs to learn a transferable network which is resource-consuming. Moreover, most of the mechanisms \cite{cai2017active,gao2018active,abel2019regional} overconfidently rely on the information of node embeddings which is known to be misleading especially when model is under-trained during AL \cite{lewis1994sequential, ducoffe2018adversarial}. In response to these challenges, we design a hybrid graph uncertainty reduction strategy to approximate the model performance measure of interest to select informative nodes. In particular, we introduce label propagation (LP) \cite{long2008graph} to facilitate the training oracle without retraining the model when estimating the node informativeness. Label propagation~\cite{ji2012variance, zhou2003learning, long2008graph, zhu2003semi} is a typical type of graph-based semi-supervised learning method where the model speedily diffuses the known label information to the graph based on node interactions. In addition, we extensively exploit the graph explicit information from different perspectives to reduce model bias and narrow down the nodes selection pool for saving computation costs.

To this end, we focus on realistic AL cases where only very few labeled data are available to obtain in our study, and propose a multi-stage AL framework called \ap{} for graph representation learning. In \ap{}, we make most of the graph topological information in the beginning stage to sample centralized nodes as our candidates. It not only benefits from the explicit graph knowledge but also narrows down the candidates for computation efficiency. Then, we design a hybrid graph uncertainty reduction metric to approximately measure the implicit contribution of candidate nodes to the model performance. Since it is unrealistic to retrain the model every time to estimate the information gain for each new candidate node during the query procedure, we use LP \cite{zhou2003learning} to simulate the GNN training in a feasible way and quickly assess the node informativeness in terms of the model's performance enhancement. We conduct comprehensive experiments to evaluate the performance of our framework on benchmark datasets and demonstrate its effectiveness.

    
\section{Problem Statement}

In this work, we focus on AL with limited labeling budget for node classification task where only a small number of nodes are available to annotate. Following the footsteps of \cite{cai2017active, gao2018active, hu2020graph, abel2019regional}, we adopt a graph convolutional network (GCN) \cite{kipf2016semi} in our semi-supervised active graph learning framework. 

Without loss of generality, we denote an undirected and unweighted graph $G$ to be $G = (V, E, X)$, where $V$ ($n = |V|$) is the set of graph nodes, $E$ is the set of edges specifying relationships, and $X \in \mathbb{R}^{n \times d}$ is feature matrix. Nodes $V$ are divided into labeled node set $V_{l}$ ($n_{l} = |V_{l}|$) and unlabeled node set $V_{u}$ ($n_{u} = |V_{u}|$). The GCN model is trained to perform the node classification through AL. In the training oracle, the $V_{l}$ is randomly initialized with a small set of labeled nodes. The model starts with the regular training on the set $V_{l}$ in each iteration. Given the model trained in the iteration $t_{i}$, a pool of unlabeled nodes $V_{u}$ and labeling budget $B$, our framework proceeds to select one unlabeled node from $V_{u}$ to annotate, and update the $V_{l}$ to the next training oracle $t_{i+1}$. The node query procedure stops until $n_{l}$ reaches the budget $B$ and whole AL framework ends until finishing the training. Therefore, the objective of this work is to optimize the GCN's performance by integrating the designed AL query strategy to choose $V_{l}$. The loss function is defined as the cross-entropy error over all labeled nodes,
\begin{equation}\label{eq:gnnloss}
W^{*} = \underset{W}{\arg\!\min}\ l(Z_{l}, y_{l}) + \lambda\|W\|_{2}^{2} 
\end{equation}
where $W$ is the trainable weight matrix, $y_{l}$ is the indices of training nodes $V_{l}$, and $\lambda$ is the hyperparameter. Notely, $V_{l}$ comes from two parts: the initialized labeled nodes $V_{init}$; and the queried set via our node query strategy $V_{query}$.



\section{Proposed Model}

\begin{figure}[t]
\vspace{-0.3cm}
    \centering
    \includegraphics[width=\linewidth]{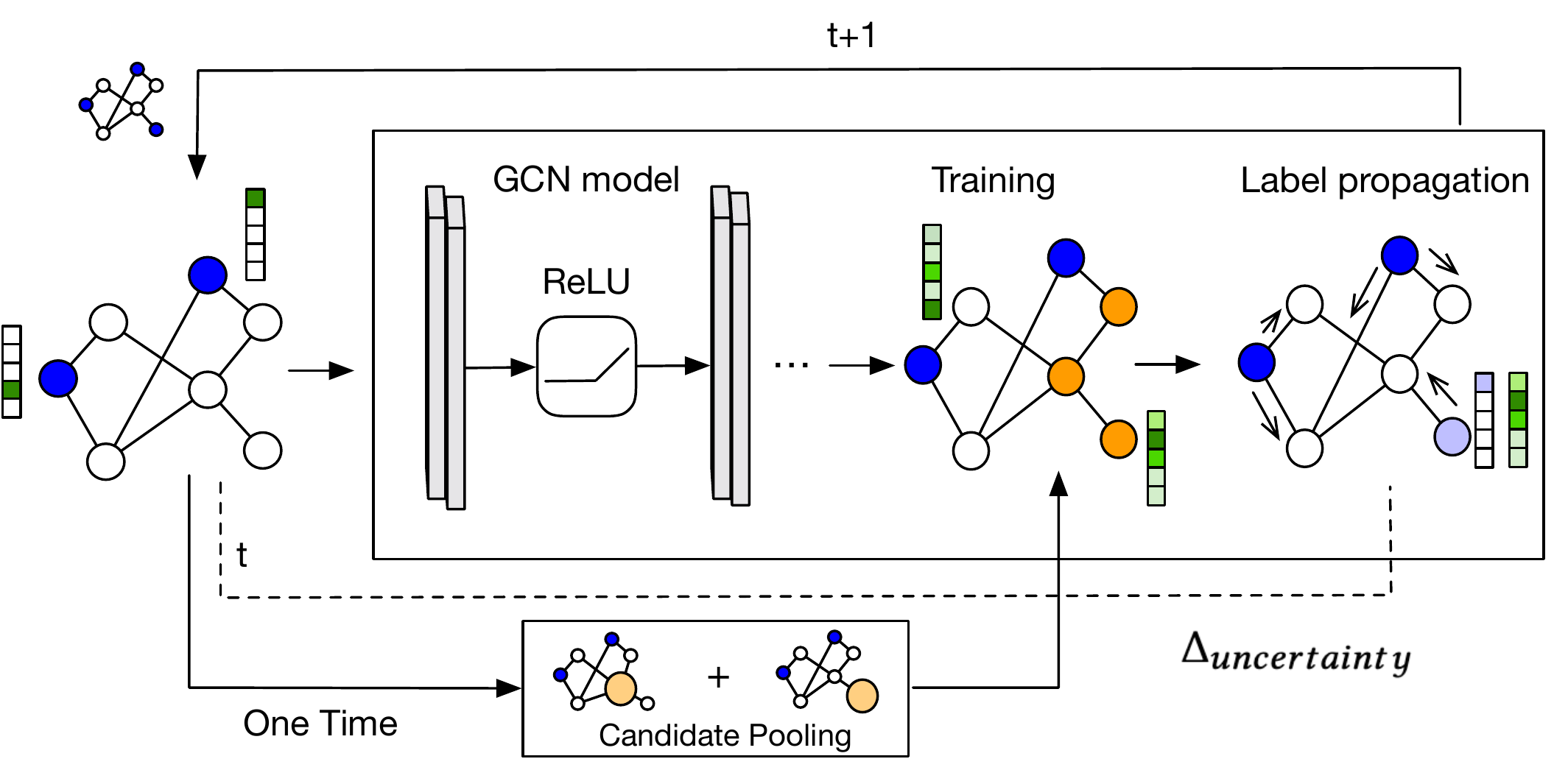}
    \caption{The \ap{} conducts one-time candidate pooling to filter representative candidates. In each iteration, the GCN model trains on the updated labeled set, then LP is performed to approximately estimate the hybrid graph uncertainty reduction of each unlabeled candidate and select one to annotate. When reaching the labeling budget, the model continuously completes the regular training.}
    \label{fig:workflow}
    \vspace{-0.4cm}
\end{figure}

Figure~\ref{fig:workflow} shows our \ap{} framework which consists of three major modules:
(1) GCN Training (2) Candidate Pooling and (3) Hybrid uncertainty reduction through LP. The details are explained in the following sections.

\subsection{Graph Uncertainty Reduction}

In the beginning iteration, a base model $G$ is trained on the initial labeled set. Then the framework proceeds with annotating one more selected node each time and continuously train the model on updated labeled set $V_{l}$. As one of the most common query criteria, uncertainty is widely used to measure the node informativeness \cite{cai2017active, gao2018active}. The nodes with high uncertainty are considered to be more informative. Specifically, it is usually obtained via computing the entropy of node embedding results. 
However, the GCN model in AL tends to be highly under-trained and unreliable that might be overconfident about the nodes it knows nothing about \cite{settles2009active}. Relying on singular node embedding to compute uncertainty can riskily introduce model biases \cite{cai2017active, gao2018active}. To avoid that, we first compute the graph uncertainty to indicate the model's uncertain status by adding up the entropy of all node embeddings during training,
\begin{equation}
    H_{G} = \sum_{v_{i}\in V}\sum_{k}^{K}(-p_{v_{ik}}\log p_{v_{ik}})
\end{equation}
where $p_{v_{ik}}$ is the probability result of the classifier for node $v_{i}$ being classified as $k$, and $K$ is the number of classes. Then we compute the graph uncertainty reduction $\Delta H_G$ in two sequential training stages $[t_i, t_{(i+1)}]$ to approximately estimate the information gain of model by selecting different nodes for training. 

However, there are two main challenges in such designs. First, measuring all unlabeled nodes to estimate the information gain is inapplicable as the size of unlabeled data pool can be very large. Moreover, it is infeasible to retrain the model step ahead with every new labeled candidate to calculate the graph uncertainty reduction. Second, the entropy overemphasizes implicit information from the embedding space which is inaccurate and untrustworthy in model's training stage, while the graph's explicit structural information is crucial for graph learning and should not be underrated. They can play a vital role in ruling out non-important nodes in model training and guide our node selections. Therefore, we delicately design \textit{Candidate Pooling} and \textit{LP-based Hybrid Uncertainty Reduction} to advance our method.

\subsection{Candidate Pooling}
In this step, we aim to design a pooling strategy to leverage more accurate explicit graph information to sample out more representative nodes at the beginning of node selection. By filtering out a large body of unlabeled candidates, we can also extremely reduce the following computation cost. Specifically, we utilize two types of informativeness measurements to sample nodes.


\noindent\textbf{Degree informativeness:} node locates at dense regions is considered to have more impact. Especially for GCN, node with more connections tends to have broader impact in information aggregation and thus are viewed to be more valuable to guide model training. In this step, we straightforwardly calculate the degree of each node as $\sum A_{ij}$, where $A$ is the adjacency matrix of graph, $A_{ij}=1$ if node $v_{i}$ connects with node $v_{j}$, otherwise $0$.

\noindent\textbf{PageRank informativeness:} it focus on the number and the quality of node links within graph. PageRank \cite{page1999pagerank} is a well-known metric in assessing node centrality in graph \cite{cai2017active, gao2018active}. Unlike the degree metric, it weighs more emphasis on node interactions. The node with low degree can possibly have high PageRank value as it may link with important nodes. It helps us explore the intrinsic graph structural information to measure the node information comprehensively. Specifically, it is calculated as follows,
\begin{equation}
    PageRank(v_{i}) = \frac{1 - \beta}{n} + \beta\sum_{j}A_{ij}\frac{\phi_{link}(v_{j})}{degree_{j}}
\end{equation}
where $\beta$ is the damping parameter, $degree_{j}$ indicates the degree of node $v_{j}$ in graph. We simply add the two normalized metric results as the pooling score and choose the top ones as final candidates for future node querying process. 


\subsection{Hybrid uncertainty Reduction through LP}
Once we restrict our candidate pool to a reasonable size, we estimate the performance improvement that each candidate can bring to the model and choose the best one to annotate in each iteration. To simulate the GCN training, we choose LP \cite{zhou2003learning} to efficiently compute the estimated graph uncertainty. LP is a traditional way of learning from labeled and unlabeled data. It focus on the known labels and can quickly spread the label information to its neighbors. In our case, we define the symmetrically normalized
adjacency matrix $\hat{A}$ as our label spreading matrix in LP. The label propagation rule is defined as,
\begin{equation}
    F(t+1) = \alpha\hat{A}F(t) + (1 - \alpha)y_{l}
\end{equation}
where $\alpha$ is the parameter in $(0,1)$, which controls the percentage of information that node received from its neighbors and its initial label information. For each candidate $v_{i}$, we annotate it with all possible classes $K$ and use LP to propagate the label information until converge for each class. Then we compute the approximated graph uncertainty reduction for as follows
\begin{equation}
    \Delta H(V_{l},v_{ik}) = H_{G}(V_{l}) - H_{G}^{lp}(LP(V_{l}, v_{ik}))
\end{equation}
where $H_{G}^{lp}(LP(V_{l}, v_{ik})$ is the graph uncertainty after label propagation with the current labled set $V_{l}$ and labeled candidate node $v_{i}$ which is annotated as class $k$. 

For each candidate, we aim to label it all possible classes to compute the expected graph uncertainty reduction. However, different classses do not share the same annotation probabilities. Take binary classification as an example, if a node that tends to be classified as label $1$ is labeled as $0$, it won't contribute the performance gain but hurt the model performance. To handle this and take advantage of the implicit node embedding information, we introduce nodes' predicted class probabilities to our query criteria and design a hybrid graph uncertainty reduction as follows,
\vspace{-0.1cm}
\begin{equation}
\vspace{-0.1cm}
    \phi_{v_{i}} = \sum_{k\in K}p_{ik}\Delta H(V_{l}, v_{ik})
\end{equation}
where $p_{ik}$ is the probability of node $v_{i}$ being class $k$ from current GCN model. According to the calculated hybrid graph uncertainty reduction, we select the node with largest value to annotate and add it to the labeled set for further training.
\section{Experiments}

\subsection{Experimental Setup}

\textbf{Datasets.} We evaluate the performance of our proposed framework on three benchmark datasets: Cora, Citeseer, and Pubmed \cite{sen2008collective}. In these citation networks, each node represents a publication document and each edge indicates the citation relations between two works. Table~\ref{tab:dataset} summarizes their basic properties.

\begin{table}[t]
    \centering 
    \caption{\label{tab:dataset}Statistics of three datasets}
    \vspace{-0.4cm}
    \tabcolsep=3pt
    \begin{tabular}{cccccc}
    \toprule
        \textbf{Dataset} &\textbf{\#Nodes} &\textbf{\#Edges} &\textbf{\#Classes} &\textbf{Features} &\textbf{Label rate}\\
        \midrule
         Cora & 2,798 & 5,429 & 7 & 1,433 & 0.0125\\
         Citeseer & 3,327 & 4,732 & 6 & 3,703 & 0.0090\\
         Pubmed & 19,717 & 44,338 & 3 & 500 & 0.0008\\
         \bottomrule
    \end{tabular}
    \vspace{-0.4cm}
\end{table}

\noindent\textbf{Baselines.} In our study, we choose six state-of-the-art GNN-based AL methods to be our baselines. (1) \textit{AGE} \cite{cai2017active}: It linearly combines three metrics, and then labelizes the node with highest score; (2) \textit{ABRMAB} \cite{gao2018active}: It upgrades AGE by learning the combination weights of metrics using an exponential MAB updating rule; (3) \textit{Coreset} \cite{sener2017active}: It performs a K-Center clustering over the node 
embeddings and tends to annotate those center nodes.
(4) \textit{Degree}: It iteratively annotates the node with the largest degree in graph; (5) \textit{Random} \cite{hu2020graph}: It randomly chooses nodes from the unlabeled set to annotate; (6) \textit{Entropy} \cite{hu2020graph}: It computes the entropy of nodes' predicted probability vector and annotate the one with the highest value.

    

\begin{table*}[t]
    \centering
    \small
    \caption{Comparisons with State-of-the-art GNN-based AL methods(\%)}
    \vspace{-0.2cm}
    \label{tab:comparison}
    \tabcolsep=2.5pt
    \begin{tabular}{lccccccccccccc}
    \toprule
        \multirow{2}{*}{}
         & \multicolumn{2}{c}{AGE}
         & \multicolumn{2}{c}{ABRMAB}
         & \multicolumn{2}{c}{Coreset}
         & \multicolumn{2}{c}{Random}
         & \multicolumn{2}{c}{Entropy}
         & \multicolumn{2}{c}{\ap{}} \\
         \cmidrule{2-13}
         & MacroF1 & MicroF1& MacroF1 & MicroF1& MacroF1 & MicroF1& MacroF1 & MicroF1& MacroF1 & MicroF1& MacroF1 & MicroF1 \\
         \midrule
         Cora & 64.44$\pm$3.7 & 69.58$\pm$2.8 & 58.96$\pm$9.3 & 65.47$\pm$6.6 & 51.16$\pm$7.8 &60.46$\pm$7.3 &58.85$\pm$7.1 &65.11$\pm$5.1 &58.02$\pm$7.9 &64.80$\pm$5.7 & \textbf{69.33$\pm$3.3} &\textbf{73.88$\pm$2.5}\\
         CiteSeer & 51.01$\pm$10.2 &59.32$\pm$9.4 &36.63$\pm$14.9 &43.76$\pm$13.0 &23.25$\pm$12.3 &35.57$\pm$10.8 &32.71$\pm$17.9&42.15$\pm$16.1 &29.24$\pm$10.8 &39.84$\pm$9.2 &\textbf{55.03$\pm$3.8}&\textbf{62.58$\pm$2.9}\\
         Pubmed & 65.98$\pm$9.5&\textbf{72.18$\pm$3.8} &57.61$\pm$16.6 &62.97$\pm$11.2 &54.56$\pm$16.1 &60.92$\pm$9.5 &54.62$\pm$11.7 &61.28$\pm$6.4&62.43$\pm$10.6 &65.64$\pm$7.8&\textbf{67.55$\pm$6.2} &71.05$\pm$5.3\\
         \bottomrule
    \end{tabular}
    \vspace{-0.3cm}
\end{table*}

\noindent\textbf{Parameter settings.} In our experiments, we set the training epoch as $300$. We use $1,000$ labeled nodes as testing set, and sample $500$ nodes from the non-testing nodes for validation. We run $10$ times random sampling to generate different validation sets. For each setting, we repeat $20$ times experiments for evaluation. Therefore, each evaluation is conducted $200$ times and we obtain their average values as final results. In each dataset, we set the number of initialized labeled nodes to be $L_{init} = 1$ per class and restrict the maximum number of labeled nodes to be $L_{max} = 5$ per class. The labeling budget $B=L_{max}-L_{init}$ for each class. The label rate in Table~\ref{tab:dataset} indicates the percentage of labeled nodes over the total number of nodes. In the candidate pooling, we set the pool size as $2\times L_{max}$ per class. As evaluation metrics, we adopt Macro-F1 and Micro-F1 \cite{perozzi2014deepwalk} to measure the node classification performance.



\vspace{-0.1cm}
\subsection{Evaluation}
\textbf{Effectiveness.} In this section, we compare \ap{} against baseline strategies and comprehensively analyze its performance over three datasets under different settings.

\subsubsection{Comparisons with Baselines}

In this part, we compare our method with state-of-the-art strategies including AGE \cite{cai2017active}, ABRMAB \cite{gao2018active}, Coreset \cite{sener2017active}, Random \cite{hu2020graph} and Entropy\cite{hu2020graph} over three different datasets. Table \ref{tab:comparison} illustrates the evaluation results in a format of classification accuracy and its standard deviation. From the table, we surprisingly find that the Random can achieve comparable results with the Entropy which further verifies the randomness and unreliability of graph embeddings during AL, particularly for those training cases with few labeled data. From the results, we can also learn that our method outperforms all the baselines on Cora and Citeseer. To be specific, \ap{} improves the classification accuracy on Cora to a scale of (4.89\%-18.17\%) on Macro-F1 and (4.30\%-13.42\%) on Micro-F1, and achieves an obvious performance increase on Macro-F1 (4.02\%-31.78\%) and Micro-F1 (3.26\%-27.01\%) for Citeseer. For Pubmed, \ap{} obtains higher results than the majority of baselines except that it slightly underperforms AGE on Micro-F1. However, our method turns out to have better generalization than AGE over all datasets as it obtains more stabilized results (smaller invariants on average).  

\vspace{-0.1cm}
\subsubsection{Effectiveness on Different Labeling Budgets} 

\begin{figure}[t]
    \centering
    \includegraphics[width=\linewidth]{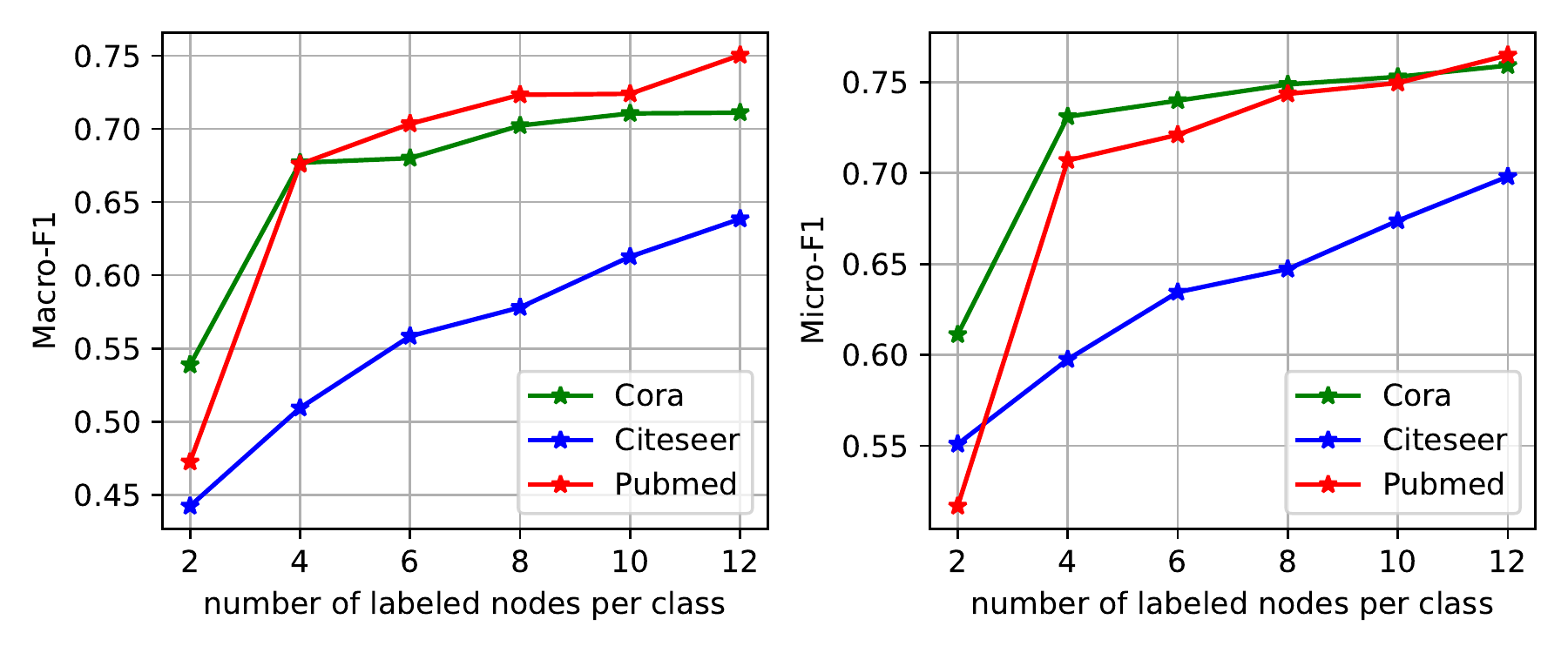}
    \vspace{-0.4cm}
    \caption{The Macro-F1 and Micro-F1 performance comparisons under different number of labeled nodes per class in three datasets.}
    \label{fig:parameter}
    \vspace{-0.3cm}
\end{figure}

In this experiment, we assess the model's performance in \ap{} with different labeling budgets for each dataset. Specifically, we evaluate our AL-based GCN training framework with $L_{max} \in \{2, 4, 6, 8, 10, 12\}$ for each class of the dataset respectively, while the other parameters retain the same setting. The experimental results are presented in Figure~\ref{fig:parameter}. We can find that with the training size increases, the model performance improves gradually as more node information is exploited. But the upward speed slows down since the newly-added nodes do not provide as much benefit as prior ones to the model information gains. The overall performance varies in different datasets due to various underlying distributions. Note that our method achieves intriguing results for Pubmed on both Macro-F1 and Micro-F1 even it has the smallest label rate.

\subsection{Ablation Study}

In this section, we conduct the ablation study to further investigate how different components contribute to the performance of our framework design. From its composition, we continuously add different components and formulate three types of workflows: (1) GCN: 
regular GCN training with randomly selected nodes for annotation; (2) GCN+Pool: annotate the node with the highest metric score from pooling procedure; (3) GCN+Pool+LP: the complete design of our method. To validate the benefit of leveraging graph explicit information in our pooling step, we add two variants: (4) GCN+LP: replace the original pooling with random sampling; (5) GCN+embedding+LP: followed by \cite{cai2017active}, we utilize Kmeans to compute the information density based on node embeddings as the pooling criteria. The results are reported in Table \ref{tab:ablation}.

\begin{table}[t]
    \centering
    \small
    \caption{Evaluation on \ap{} Components (\%)}
    \vspace{-0.3cm}
    \label{tab:ablation}
    \tabcolsep=1.5pt
    \begin{tabular}{lccccccc}
    \toprule
        \multirow{2}{*}{}
         & \multicolumn{2}{c}{Cora}
         & \multicolumn{2}{c}{Citeseer}
         & \multicolumn{2}{c}{Pubmed} \\
         \cmidrule{2-7}
         & MacroF1 & MicroF1& MacroF1 & MicroF1& MacroF1 & MicroF1 \\
         \midrule
         GCN & 58.85&65.11&32.71&42.15&54.62&61.28\\
         GCN+Pool & 62.47&66.65&49.10&55.20&61.83&68.68\\
         GCN+LP & 59.18& 65.29&38.68&47.42&60.42 & 62.55\\
         GCN+emb.+LP &44.86&55.42&34.49&43.25&58.48&60.36 \\
         GCN+Pool+LP &\textbf{69.33}&\textbf{73.88}&\textbf{55.03}&\textbf{62.58} &\textbf{67.55} &\textbf{71.05}\\
         \bottomrule
    \end{tabular}
    \vspace{-0.4cm}
\end{table}

From the results, we can see that the pooling step plays an important role as it significantly upgrades the GCN performance by 16.39\%. It proves that the explicit information we extract from graph can accurately lead us to find those more informative nodes in the early stage. In contrast, over-relying on embeddings from under-trained model can even degrade the results as the predicted results can be biased, e.g., $GCN+embed.+LP$ underperforms $GCN+LP$. The LP component can slightly improve the GCN training as it provides more insight towards information gain than random selections, our complete design can further advance the model performance to a high level. These observations reaffirm the superiority of our design on AL over GCNs within very few labeled data.

\section{Conclusion}
In this work, we design a practical AL framework for GNNs named \ap{}. It calculates the implicit information gain from different nodes by utilizing a hybrid uncertainty reduction scheme to automatically select representative nodes and facilitate active graph learning. Moreover, we extract explicit graph structural information for node querying guidance to effectively reduce model uncertainty and expedite the training. We conduct extensive experiments and the ablation study to unveil the insight of integrating different sources of graph information for AL-based GNN training under very few labeled nodes. The state-of-the-art results validate its effectiveness and superiority against baselines.

\bibliographystyle{ACM-Reference-Format}
\bibliography{main}


\end{document}